# ENHANCING PASHTO TEXT CLASSIFICATION USING LANGUAGE PROCESSING TECHNIQUES FOR SINGLE AND MULTI-LABEL ANALYSIS

Mursal Dawodi and Jawid Ahmad Baktash

LIA, Avignon University, Avignon, France
mursal.dawodi@gmail.com
jawid.baktash1989@gmail.com

## ABSTRACT

*Text classification has become a crucial task in various fields, leading to a significant amount of research on developing automated text classification systems for national and international languages. However, there is a growing need for automated text classification systems that can handle local languages. This study aims to establish an automated classification system for Pashto text. To achieve this goal, we constructed a dataset of Pashto documents and applied various models, including statistical and neural machine learning models such as DistilBERT-base-multilingual-cased, Multilayer Perceptron, Support Vector Machine, K Nearest Neighbor, decision tree, Gaussian naïve Bayes, multinomial naïve Bayes, random forest, and logistic regression, to identify the most effective approach. We also evaluated two different feature extraction methods, bag of words and Term Frequency Inverse Document Frequency. The study achieved an average testing accuracy rate of 94% using the MLP classification algorithm and TFIDF feature extraction method in single-label multiclass classification. Similarly, MLP+TFIDF yielded the best results, with an F1-measure of 0.81. Furthermore, the use of pre-trained language representation models, such as DistilBERT, showed promising results for Pashto text classification; however, the study highlights the importance of developing a specific tokenizer for a particular language to achieve reasonable results.*

## 1. INTRODUCTION

The evolution of technology instigated the existence of an overwhelming number of electronic documents therefore text mining becomes a crucial task. Many businesses and individuals use machine learning techniques to classify documents accurately and quickly. On the other hand, more than 80% of organization information is in electronic format including news, email, data about users, reports, etc. (Raghavan, 2004). Text mining attracted the attention of researchers to automatically figure out the patterns of millions of electronic texts. Among other opportunities, this provides a facility for users to discover the most desirable text/document.

Pashto is a resource poor language and the unavailability of standard, public, free of cost datasets of text documents is a major obstacle for Pashto's document classification. Automatic text document classification and comparatively analyse the performance of different models are the main gaps in Pashto text mining. This research is the first attempt to classify Pashto documents into eight classes including Sport, History, Health, Scientific, Cultural, Economic, Political, and Technology. Besides, this is the initial and novel work on Pashto text multi-label classification. The main contributions of this research are to:

- Designing two Pashto document dataset and make them publicly and free of cost available in the future.

- Compare the performance of 32 distinct models on Pashto text single label and multilabel classification.

- Evaluate the performance of standard pre-trained language representation model (DistilBERT) on Pashto language processing.

## 2. PASHTO LANGUAGE

Pashto is an Iranian language, belonging to the Indo-European language family, spoken natively by over seven million Afghans, more than seven million Pakistanis, and approximately 5,000 Iranians [14]. Pashtuns, individuals who have Pashto as their mother tongue, primarily reside in the southern regions of Afghanistan and northern regions of Pakistan. This language is characterized by three distinct dialects that vary based the geographic location of the native Pashtun residents. The diversity of dialects can even affect the spelling of Pashto text, as some speakers may pronounce the sound "sh" like the Greek letter "x" or the German "ch" instead of the English "sh" [14]. Moreover, there is no standard transliteration for rendering Pashto text in the Roman alphabet, hence both Pashto and Pashtu are considered correct spelling forms [14]. Despite this, official recommendations regarding Pashto writing and speaking can be found. However, there are no standard rules for writing and pronunciation, and as such, authors may write the same word in multiple ways, and speakers may pronounce them in various ways [14]. The Pashto alphabet is similar to Arabic and Persian with some additional characters. Figure 1 illustrates the alphabet representation in the Pashto language.

Figure 1. Pashto Alphabet

Pashto differentiate nouns based on genders and distinguishes the form of verbs and pronouns for masculine and feminine nouns, as an example, دا د هغې مور ده (daa de haghe mor da) means she is her mother and دا د هغه پلار دی (daa de hagha pelar de) indicates he is his father. Morphemes like plural morphemes in Pashto added another challenge to this language [9], e.g. the plural form of زوی (son) is زامن (zaamen, sons) while کتابونه (ketaboona, books) is the plural form of كتاب (ketab, book) and the plural form of انجلۍ (enjeley, girl) is انجوني (anjoone, girls). Besides, news, articles, and other online and offline scripts are not written/typed by Pashto native speakers hence the probability of grammar and spelling error is high [14]. Additionally, grammar in Pashto is not as traditional as other Indo-European languages. Although nowadays several Pashto grammar books are published. Still, they have contradicted each other in some parts [14]. Furthermore, other languages spoken in the vicinity of Pashtun areas have major influences on this language that caused arriving of foreign words in Pashto for instance. some Pashtuns combine Urdu or Dari words with Pashto while speaking or in their written text.

## 3. RELATED WORKS

Many studies on document classification have already been conducted in international and western languages. As a recent work in text document classification, Gutiérrez et al. [7] developed a COVID 19 document classification system. They compared several algorithms including SVM, LSTM, LSTMreg, Logistic Regression, XML-CNN, KimCNN, BERTbase, BERTlarg, Longformer, and BioBERT. The best performance was achieved by BioBERT with an accuracy of 75.2% and a micro-F1-measure of 0.862 on the test set.

In recent years some researchers started to work on document classification in Asian and local languages. Ghasemi S. and Jadidinejad A.H. [6] used character level convolutional neural network to classify Persian documents. They obtained 49% accuracy which was much higher compared to the results of Naïve Bayes and SVM. Similarly, Baygin M. [2] used the Naïve Bayes method and n-gram features to classify documents in Turkey into economic, health, sports, political, and magazine newsgroups. They performed their proposed model on 1150 documents written in Turkey. The best performance was achieved by the 3-gram technique with 97% accuracy on sport, politics, and health documents, 98% on a magazine, and 94% on economic documents.

Similarly, Pervez et al. [12] obtained impressive results using a single layer convolutional neural network with different kernel sizes to classify Urdu documents. They evaluated the model on three different Urdu datasets including NPUU, naïve, and COUNTER. NPUU corpus consists of sport, economics, environment, business, crime, politics, and science and technology Urdu documents. Likewise, naïve contains Urdu documents related to sports, politics, entertainment, and economic. Finally, the main document classes in COUNTER dataset are business, showbiz, sports, foreign, and national. Consequently, they obtained 95.1%, 91.4%, and 90.1% accuracy on naïve, the COUNTER, and NPUU datasets, respectively. Pal, Patel, and Biraj [11]. Automatic Multiclass Document Classification of Hindi Poems using Machine Learning Techniques. 2020 International Conference for Emerging Technology (INCET), 1–5. https://doi.org/10.1109/INCET49848.2020.9154001 categorized Indi poem documents into three classes romance, heroic, and pity according to the purpose of the poem. They evaluated several machine learning techniques. The maximum accuracy 56%, 54%, 44%, 64%, and 52% using Random Forest, KNN, Decision Tree, Multinomial Naïve Bayes, SVM, and Gausian Naïve Bayes.

Some researches were conducted in the context of multi-label classification of articles, recently. As a similar work [4], constructed two separate large corpora for single label and multi-label Arabic news categorization. They evaluated the performance of several deep learning algorithms on classifying Arabic articles. Finally, the best performance of 96.94% accuracy and 88.68% overall accuracy using attention-GRU in the context of single and multi-label classification process, respectively. Similarly, Al-Salemi et al. [1] introduced a new Arabic multi-label benchmark dataset named "RTANews". Next, they examine the performance of four problem transformation-based approaches, including Binary Relevance, Classifier Chains, Calibrated Ranking by Pairwise Comparison, and Label Powerset, and five algorithm adaptation-based techniques, including Multi-label K-Nearest Neighbors, Instance-Based Learning by Logistic Regression Multi-label, Binary Relevance KNN, and RFBoost. As a result, the transformation approaches were performed better with SVM as a base classifier, and the best performance was achieved with RFBoost method. Likewise, Qadi et al. [13] established an Arabic multi-label dataset with four main categories: Business, Sports, Technology, and the Middle East. They utilized Logistic Regression, Nearest Centroid, DT, SVM, KNN, XGBoost, Random Forest Classifier, Multinomial, Ada-Boost, and MLP to determine relevant labels for Arabic news. They claimed that SVM with 97.6% accuracy outperformed other methods.

Recent studies utilized transformer based pre-trained models. Tokgoz et al. [15] used BERT and DistilBERT with different tokenizers including Turkish tokenizer to classify news in Turkish language. DistilBERT with Turkish tokenizer obtained the best performance with 97.4% accuracy. Similarly, the study by Gutiérrez et al. [7] compared the performance of traditional machine learning methods, convolutional neural models, and pretrained language models on COVID 19 document classification. As a result, the reasonable accuracy of 75.2% and 74.4% achieved with BioBERT and BERT large respectively. Correspondingly, Farahani et al. [5] developed a monolingual transformer-based model for Persian language and evaluated the proposed model in several datasets. Finally, the investigation declares that the ParsBBERT outperforming both multilingual BERT and other prior works in Persian down-stream NLP tasks such as Sentiment Analysis, Text Classification and Named Entity Recognition. Dai and Liu [3] used BERT model for multilabel classification of Chinese Judicial documents.

As of history, there is no documented classification for the Pashto language. The only work to classify the Pashto text, which in some respects relates to our work, was done by S. Zahoor et al. [16]. They have developed a character recognition system that captures images of Pashto letters and automatically classifies them by predicting a single character.

## 4. CORPORA

In this study we constructed two datasets corresponding to multilabel and single label multiclass document classification.

### 4.1   Single Label Dataset

This research gathered 800 manuscripts from several online books, articles, and web pages to make a corpus for text document classification analysis. Subsequently, we manually assigned label/s by setting a number to every single document in relevance to the category it belongs. We collected 100 Pashto documents for each class including history, technology, sport, cultural, economic, health, political, and scientific. Besides, we increased the dataset with 475 news-related documents.

### 4.2   Multi Label Dataset

The structure of the corpus is altered compared to single-label document corpus. Each document was assigned to multiple related classes. Here, we considered the news category along with the prior labels. The average length of the documents is 3119 words where the shortest document has 232 words and the longest one includes 31740 words. Table 1 shows the total number of related documents per label in a multi-label corpus and figure 2 depicts the range of labels used for every document. Similarly, the total number of words is 3289214 in the single label and 3976976 words in multi-label datasets. The average number of labels per document is 2.5.

In the next step, the authors pre-processed the dataset by applying some spelling and grammar modification, removing any noisy and senseless symbols including non-language characters, special symbols, numeric values, and URLs. As a result, we standardized and normalized the texts within the documents.

Table 1 Total number of documents related to each label in multi-label classification

| Label | Documents |
|---|---|
| History | 1251 |
| Culture | 1276 |
| Economic | 1274 |
| Health | 1275 |
| Politic | 1263 |
| Scientific | 1270 |
| Sport | 1182 |
| Technology | 1276 |
| News | 1273 |

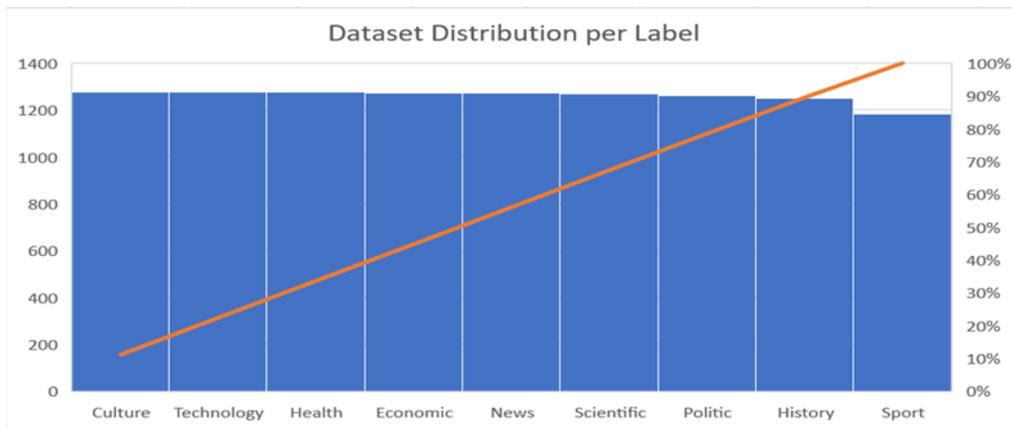

Figure 2. Distribution of each label within the multi-label dataset

## 5. MODELS

This section details the methods followed to accomplish the study for categorizing Pashto sentences and documents. We used three different types of classifiers with different feature extraction and tokenization methods.

### 5.1 Traditional Models

As mentioned in prior sections, this work observed different classifiers methods including Naïve Bayes, Multinomial Naïve Bayes, Nearest Neighbour, Random Forest, Decision Tree, SVM, Logistic Regression. This project fine-tuned the value of K finally the result shows that the optimum value for k occurs at k=5.

### 5.2 Neural Network Models

We used Multilayer Perceptron Network (MLP) as the neural network model. MLP is a neural network classifier which is a subset of machine learning consists of neurons and layers. MLP consists of several layers including one input, one output, and hidden layers. The outputs of one perceptron are fed as input to subsequent perceptron. This experiment used a single hidden layer with 20 neurons. We used backpropagation and gradient descent to provide the ability to propagate errors back to earlier layers. Moreover, we shuffled the samples to reduce noise by feeding different inputs to neurons in each iteration and as a result, make good generalizations. The activation function used in this model is Rectified Linear Unit (ReLU) function, which is a non-linear activation function, to decrease the chance of vanishing gradient. ReLU is defined in equation 1 where f(a) is considered to be zero for all negative numbers of a. Finally, we used Adam as an optimization algorithm.

$$f(a) = \max(0, a) \text{ where } a = Wx + b \quad (1)$$

### 5.3 Contextual Word Representation Models

Nowadays, contextual word representation models such as BERT and DistilBERT are used in varied tasks of NLP including text classification. BERT was initially developed by Google AI as base and large pre-trained models. The difference between two models is on the number of transformer blocks. Base model uses 12 layers encoders where the large BERT utilizes 24 layers of encoder on top of each other. Thus, the base model contains 12 self-attention heads with hidden size of 786. The maximum number of tokens handled by base model is 512. It contains some default tokens as [CLS] that marks the starting of the segment and [SEP] that differentiates the segments. DistilBERT is faster and smaller distilled version of BERT which is more suitable in NLP tasks. The architecture of DistilBERT contains 6 layers, 12 heads, and 786 dimensions.

We used BERT-base-multilingual-cased and DistilBERT-base-multilingual-cased models with the tokenizer of BERT-base-multilingual-cased and DistilBERT-base-multilingual-cased, respectively. In our model the final hidden state of the first token [CLS] demonstrates the entire sequence. An activation function classifier on the of the model is used to predict the related class of a text document.

### 6. Tokenization

Tokenization shrinks the sentences into lexicons/tokens (e.g. ['واه', 'ډیر', 'ښکلی']) using available token list. The special tokens specify the start and end of the sequence. We specified the maximum length of tokens. Hence, the extra tokens are discarded if the sequence is longer than the maximum size while extra empty tokens are added to shorter sequences.

```
                    د دوزخ لمبی دي

ID's    Input Tokens
101     [CLS]
771     د
13669   دو
11509   ##ز
16498   ##خ
16849   لم
23772   ##بی
35640   دي
102     [SEP]
```

Figure 3 Pashto Sentence Tokenization using standard DistilBERT Tokenizer

In this study we tokenized the text by using NLTK work tokenizer in the traditional and MLP models. On the other hand, the standard tokenization process was used in BERT and BistilBERT models. The procurement root of separate tokens in Pashto is a more challenging task due to morphemes and other issues in Pashto literature. Thus, this work used lexicons in their default forms. Figure 1 represents how a sentence in Pashto is tokenized into tokens each containing an ID.

## 7. Evaluation Matrices

This study evaluates the performance of different classifiers using separate feature extraction methods. We considered four metrics Precision (equation 2), Recall (equation 3), F1-measure (equation 4), and accuracy (equation 5) to analyse the outcome of different models of the first and second group used in this experiment. Precision, which is called positive predictive values, is the percentage of examples that the classifier predicts accurately from the total samples predicted for a given tag. On the other hand, Recall which is also referred to as sensitivity determines the percentage of samples that the classifier predicts for a given label from the total number of samples that should be predicted for that label. Accuracy represents the performance of the model while is referred to the percentage of texts that are predicted with the correct label. We used F1-measure to measure the average between Precision and Recall values. There are mainly four actual classes true real positive (TP), false real positive (FP), true real negative (TN), and false real negative (FN). TP and TN are the accurate predictions while FP and FN are related to imprecise estimations:

$$\text{True class} = \{TP_1, TP_2, \ldots, TP_n,\} \cup \{TP_1, TP_2, \ldots, TP_n,\}$$

$$\text{False class} = \{FP\_1, FP\_2, \ldots, FP\_n,\} \cup \{FP\_1, FP\_2, \ldots, FP\_n,\}$$

$$Precision = \llbracket TP \rrbracket\_i / (\llbracket TP \rrbracket\_i + \llbracket FP \rrbracket\_i) \quad (2)$$

$$Recall = \llbracket TP \rrbracket\_i / (\llbracket TP \rrbracket\_i + \llbracket FN \rrbracket\_i) \quad (3)$$

$$F1 - measure = 2PR/(P + R) \quad (4)$$

$$Accuracy = (\llbracket TP \rrbracket\_i + \llbracket TN \rrbracket\_i) / (\llbracket TP \rrbracket\_i + \llbracket FP \rrbracket\_i + \llbracket TN \rrbracket\_i + \llbracket FN \rrbracket\_i) \quad (5)$$

Consequently, this study computes weighted average values for Precision, Recall, and F1-measure of all classes to compare the efficiency of each technique. In multi-label classification, the weighted average evaluates metrics for all labels and calculates their averages weighted by the total number of true instances per each label. Besides, to evaluate the performance of individual algorithms in classifying documents into multiple tags, we used AUC (area under the curve) along with other criteria. AUC is a classification threshold invariant which, means that it calculates the performance of the models concerning all possible score thresholds, regardless of the importance of each threshold. It corresponds to the array of samples and classes. The probability estimates are related to the probability of the class that has a larger label per output of the classifier. To compute AUC-ROC we used the ROC_AUC_score method of Python scikit learn metrics library. Additionally, we evaluated the sample average scores for precision, recall and f1-measure. The sample average estimates metrics per instance then averages the results. For analysis of the third model, we used accuracy and loss metrics in single label classification task. On the other hand, the authors experiment the performance of third group of models on Pashto text classification by using hamming score and hamming loss. Hamming score corresponds to the portion of accurate predictions associated to the overall labels while the fraction of wrong labels to the entire number of labels indicates the hamming loss.

## 8. Results and Discussion

This study was carried out using a computer with an Intel Core i7 processor in a Windows 10 environment, with Python 3.6.9 and TensorFlow version 2.3.1 and Keras version 2.4.3 used to implement the classification models. While this study may not surpass recent related research, such as that conducted by Elnagar [4], it is a significant contribution to the field of Pashto text classification as no prior research has been done on this topic.

However, this study does have some limitations, mainly due to the underdeveloped nature of the Pashto language. There is no specific toolkit for processing Pashto, unlike Hazm for Persian and NLTK for English. The dataset used in this study was also relatively small, containing only 800 records, and only eight classes of Pashto documents were considered. Future research aims to expand the corpus and employ more hybrid algorithms to achieve better performance.

## 8.1 Multiclass Single Label Classification Using Model I and Model II

MLP with unigram feature extraction technique illustrated the best performances among others with the gained average accuracy of 94%. Besides, it obtained 0.94 as weighted average Precision, Recall, and F1-measure scores. The maximum weighted average Precision using MLP and unigram is 0.91. The obtained results are presented in Table 2 and Figures 3. Table 2 demonstrates obtained accuracy using different techniques. Similarly, figure 3 denotes the F-measure weighted average values obtained when testing distinct techniques.

Multinomial Naïve Bayes with Unigram achieved 88% accuracy while it decreased by 7% replacing Unigram with TFIDF which indicates that it performed better with Unigram text embedding technique. However, Gaussian Naïve Bayes obtained 87% accuracy using TFIDF vector representations which are 11% higher compared to Gaussian Naïve Bayes +Unigram. Even though, Gaussian Naïve Bayes has an impressive result of 0.85 as weighted Precision result, but it obtained F1-measure of only 0.77 due to its low Recall score of 0.76. In contrast to Gaussian Naïve Bayes, Decision Tree obtains 5% more accuracy using Unigram rather than TFIDF. Performance of Logistic Regression, SVM, Random Forest, and KNN with both TFIDF and Unigram are comparable with only 1% change in accuracy and 0-0.2 variation in F1-measures.

The combination of SVM and unigram represented 84% average accuracy while this value is reduced by 1% using TFIDF. Therefore, similar to several classification studies SVM performed good in Pashto text document classification. In contrast to the work by Mohtashami and Bazrafkan [10], KNN attained only 71% as average accuracy using TFIDF method that is decreased to 70% after altering the feature extraction method from TFIDF to Unigram. The least performance belongs to Decision Tree method with TFIDF technique in this comparison experiment which is only 64% accuracy. This method also has low performance (with F1-measure of 0.69) using unigram extraction method. On the other hand, the entire methods performed their worst with bigram technique as illustrated in Table 2.

Table 2 Average accuracy using different classification and feature extraction techniques.

| Technique | Feature Extraction Method | Accuracy |
|---|---|---|
| Gaussian Naïve Bayes | Unigram | 0.76 |
| | TFIDF | 0.87 |
| | Bigram | 0.72 |
| Multinomial Naïve Bayes | Unigram | 0.88 |
| | TFIDF | 0.81 |
| | Bigram | 0.78 |
| Decision Tree | Unigram | 0.69 |
| | TFIDF | 0.64 |
| | Bigram | 0.44 |
| Random Forest | Unigram | 0.82 |
| | TFIDF | 0.81 |
| | Bigram | 0.67 |
| Logistic Regression | Unigram | 0.85 |
| | TFIDF | 0.84 |
| | Bigram | 0.36 |

| SVM | Unigram | 0.83 |
|---|---|---|
|  | TFIDF | 0.84 |
|  | Bigram | 0.65 |
| K Nearest Neighbor | Unigram | 0.7 |
|  | TFIDF | 0.71 |
|  | Bigram | 0.31 |
| Multilayer Perceptron | Unigram | 0.91 |
|  | TFIDF | 0.94 |
|  | Bigram | 0.88 |

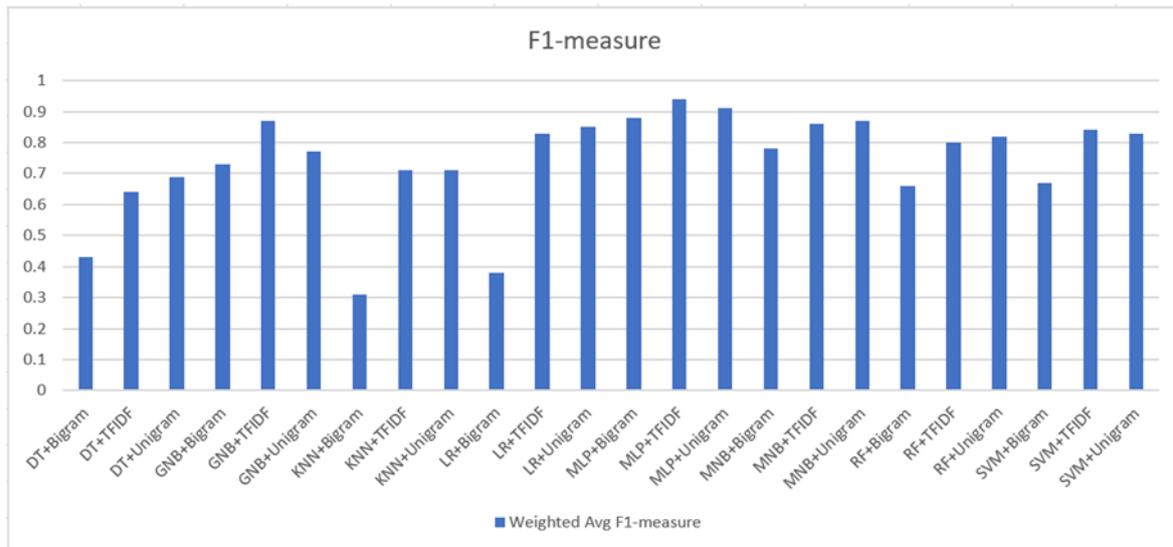

Figure 3 Weighted average F1-measure For Single Label Classification

The outcome of each model is different according to the separate class label. As an example, KNN employed unigram has 0.98 F1-measure related to History tag. However, it obtained only 0.37 for scientific documents. Similarly, all models illustrated good F1-measure for documents relevant to History except Gaussian Naïve Bayes. DT achieved high F1-measure only by predicting documents related to History. Experiments show that MLP models and the combined model of Random Forest with Unigram more accurately predicts cultural documents compared to other models. MLP with TFIDF and Gaussian Naïve Bayes with Unigram with 0.95 and 0.93 F1-measure have the most accurate Economic class predictions in this experiment. On the other hand, the implementation of Gaussian Naïve Bayes and SVM with
Unigram represents the most precise results in the context of health documents. Similarly, MLP with TFIDF obtains the highest F1-measure of 1 on predicting Politic documents. All models failed to predict scientific documents precisely except MLP + TFIDF model with F1-measure of 0.89. MLP with Unigram with F1-measure 0.98 best performed in discovering texts related to Sport class. Similarly, Random Forest and Gaussian Naïve Bayes with TFIDF came in second with F1-measure 0.95 in this era. Likewise, MLP+TFIDF and Gaussian Naïve Bayes + TFIDF models best predict texts belonging to Technology class with F1-measure 1 and 0.97 respectively.

**8.2     Multiclass Multi Label Classification Using Model I and Model II**
In multilabel classification, MLP technique illustrated the best performance similar to the single label classification with sample average F1-measure of 0.81 using TFIDF technique. Despite, the MNB+Bigram technique has the highest AUC score of 85.7% but its F1-measure is 3% lower than MLP+TFIDF. Afterward, SVM+TFIDF and SVM+ Unigram obtained sample average F1-measure of 0.74. On the other hand, according to the Precision metric the highest value of 0.86 achieved using SVM+TFIDF technique. Using any of the MNB+Bigram, MLP+TFIDF. However, Technology related documents were better detected using MLP+Bigram technique. With SVM+Unigram, MLP+Unigram algorithms the AUC scores (figure 5) are higher than 80%. The least performance obtained using LR+Bigram and LR+Trigram models.

Table 3 depicts the average accuracy obtained using different algorithms according to separate labels. Regarding separate label the highest prediction accuracy related to History, Culture, and Economics achieved by MLP+TFIDF. However, the SVM+TFIDF algorithm presented the best performance based on News, Health, and Politic label groups. On the other hand, the GNB+TFIDF predicted Scientific related documents more accurately.

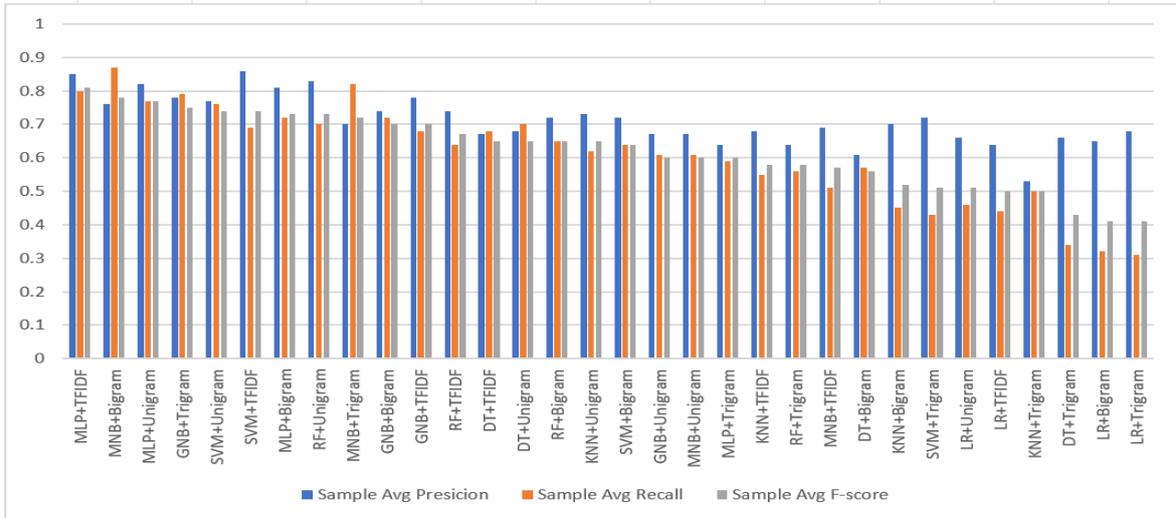

Figure 4 Sample average Precision, Recall, and F1-measure in multi-label classifier

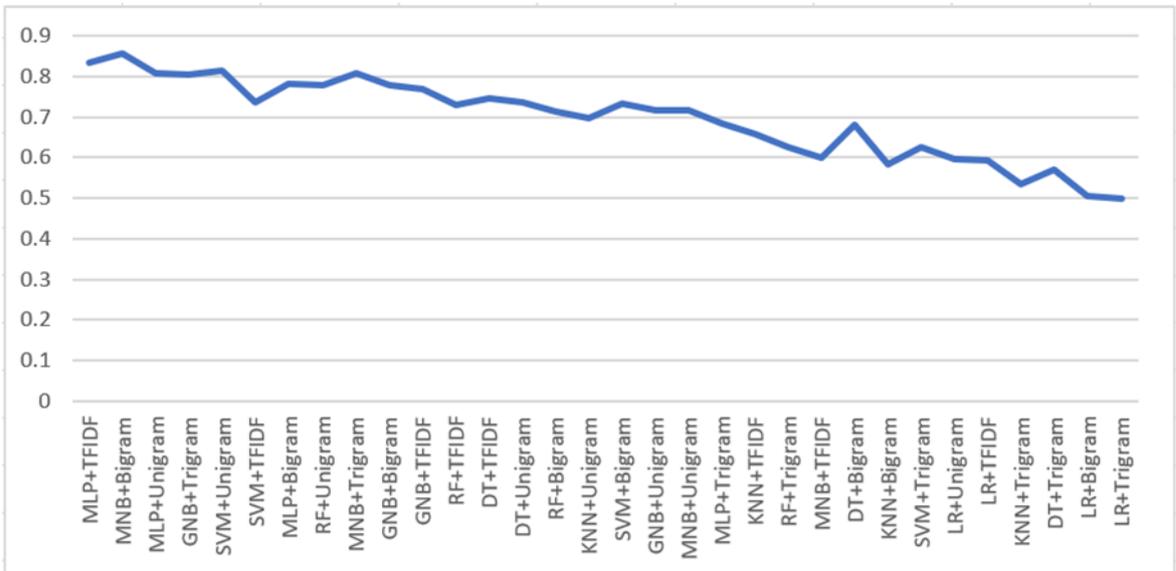

Figure 5 Compression of AUC result based on various algorithms

Table 3 Obtained accuracy related to separate labels in multi-label classification

| Technique | History | Culture | Economic | Health | Politic | Scientific | Sport | Technology | News |
|---|---|---|---|---|---|---|---|---|---|
| DT+Bigram | 0.9 | 0.65 | 0.76 | 0.82 | 0.62 | 0.81 | 0.96 | 0.91 | 0.74 |
| DT+TFIDF | 0.88 | 0.72 | 0.81 | 0.81 | 0.77 | 0.83 | 0.94 | 0.92 | 0.81 |
| DT+Unigram | 0.91 | 0.71 | 0.78 | 0.84 | 0.72 | 0.79 | 0.92 | 0.92 | 0.86 |
| GNB+Bigram | 0.87 | 0.72 | 0.83 | 0.89 | 0.82 | 0.86 | 0.98 | 0.95 | 0.84 |
| GNB+TFIDF | 0.94 | 0.78 | 0.89 | 0.92 | 0.84 | 0.94 | 0.94 | 0.93 | 0.84 |
| GNB+Trigram | 0.89 | 0.73 | 0.82 | 0.87 | 0.8 | 0.83 | 0.96 | 0.92 | 0.9 |
| GNB+Unigram | 0.93 | 0.65 | 0.86 | 0.88 | 0.8 | 0.86 | 0.96 | 0.88 | 0.8 |
| KNN+Bigram | 0.91 | 0.63 | 0.79 | 0.85 | 0.65 | 0.82 | 0.94 | 0.92 | 0.76 |
| KNN+TFIDF | 0.92 | 0.69 | 0.84 | 0.9 | 0.78 | 0.83 | 0.95 | 0.89 | 0.81 |
| KNN+Unigram | 0.93 | 0.72 | 0.83 | 0.88 | 0.8 | 0.88 | 0.96 | 0.9 | 0.85 |
| LR+Bigram | 0.89 | 0.69 | 0.84 | 0.8 | 0.58 | 0.81 | 0.94 | 0.92 | 0.64 |
| LR+TFIDF | 0.91 | 0.7 | 0.84 | 0.8 | 0.8 | 0.86 | 0.95 | 0.9 | 0.9 |
| LR+Unigram | 0.92 | 0.75 | 0.82 | 0.82 | 0.84 | 0.82 | 0.92 | 0.9 | 0.9 |
| MLP+Bigram | 0.95 | 0.78 | 0.89 | 0.89 | 0.79 | 0.9 | 0.96 | 0.96 | 0.88 |
| MLP+TFIDF | 0.96 | 0.86 | 0.92 | 0.93 | 0.89 | 0.92 | 0.96 | 0.95 | 0.94 |
| MLP+Unigram | 0.9 | 0.83 | 0.87 | 0.94 | 0.86 | 0.91 | 0.98 | 0.95 | 0.91 |
| MNB+Bigram | 0.92 | 0.78 | 0.8 | 0.87 | 0.86 | 0.89 | 0.94 | 0.91 | 0.92 |
| MNB+TFIDF | 0.9 | 0.7 | 0.84 | 0.83 | 0.86 | 0.87 | 0.93 | 0.9 | 0.9 |
| MNB+Trigram | 0.89 | 0.73 | 0.8 | 0.83 | 0.76 | 0.87 | 0.92 | 0.88 | 0.9 |
| MNB+Unigram | 0.93 | 0.65 | 0.86 | 0.88 | 0.8 | 0.86 | 0.96 | 0.88 | 0.8 |
| RF+Bigram | 0.921 | 0.733 | 0.835 | 0.898 | 0.733 | 0.858 | 0.976 | 0.909 | 0.851 |
| RF+TFIDF | 0.93 | 0.77 | 0.87 | 0.9 | 0.82 | 0.85 | 0.97 | 0.94 | 0.9 |
| RF+Unigram | 0.93 | 0.78 | 0.86 | 0.93 | 0.85 | 0.92 | 0.99 | 0.92 | 0.89 |
| SVM+Bigram | 0.91 | 0.69 | 0.84 | 0.86 | 0.74 | 0.83 | 0.94 | 0.94 | 0.85 |
| SVM+TFIDF | 0.91 | 0.8 | 0.88 | 0.94 | 0.91 | 0.88 | 0.97 | 0.93 | 0.95 |
| SVM+Trigram | 0.96 | 0.7 | 0.83 | 0.85 | 0.61 | 0.8 | 0.95 | 0.92 | 0.75 |
| SVM+Trigram | 0.96 | 0.7 | 0.83 | 0.85 | 0.61 | 0.8 | 0.95 | 0.92 | 0.75 |
| SVM+Unigram | 0.92 | 0.76 | 0.87 | 0.92 | 0.83 | 0.89 | 0.98 | 0.94 | 0.86 |
| DT+Trigram | 0.93 | 0.61 | 0.81 | 0.85 | 0.6 | 0.81 | 0.97 | 0.92 | 0.68 |
| LR+Tri | 0.95 | 0.68 | 0.82 | 0.81 | 0.58 | 0.77 | 0.9 | 0.9 | 0.67 |
| KNN+Trigram | 0.91 | 0.67 | 0.87 | 0.26 | 0.58 | 0.81 | 0.91 | 0.93 | 0.68 |
| MLP+Trigram | 0.98 | 0.68 | 0.82 | 0.83 | 0.66 | 0.8 | 0.95 | 0.92 | 0.76 |
| RF+Trigram | 0.91 | 0.63 | 0.81 | 0.85 | 0.58 | 0.87 | 0.94 | 0.93 | 0.8 |

Similarly, RF+Unigram more precisely determined sport documents. Table 3 represents the weighted average Precision, Recall, F1-measure, and Support metrics corresponding to multi-label Pashto article classification. As one can see the MLP+TFIDF technique achieves the highest weighted average Precision.

As represented in table 5, some diverse methods have different performances with variant feature extraction techniques. For example, the GNB works the best using trigram method, however, SVM outperformed with unigram. Similarly, the combination of MLP with TFIDF and MNB along with bigram performance more precisely based on this study.

MLP+TFIDF technique achieves the highest weighted average Precision. Some diverse methods have different performances with variant feature extraction techniques. For example, the GNB works the best using trigram method, however, SVM outperformed with unigram. Similarly, the combination of MLP with TFIDF and MNB along with bigram performance more precisely based on this study.

The multi-label classification models predict the exact tags for an article. Figures 6 and 7 demonstrate three news articles from the BBC Pashto News website and the predicted labels. The true labels for figure 6 are political, economic, and news. Similarly, the news article represented in figure 7 is related to sport and the last figure is the news about COVID-19 and flights between Saudi Arabia and some other countries. Fortunately, our model predicts all tags accurately.

Table 4 Weighted average Precision, Recall, F1-measure, and Support related to multi-label classification.

| Technique | Weighted Average Precision | Weighted Average Recall | Weighted Average F1-measure | Weighted Average Support |
|---|---|---|---|---|
| MLP+TFIDF | 0.89 | 0.79 | 0.84 | 533 |
| MLP+Unigram | 0.84 | 0.77 | 0.79 | 530 |
| MNB+Bigram | 0.74 | 0.87 | 0.79 | 543 |
| SVM+Unigram | 0.78 | 0.75 | 0.77 | 551 |
| MLP+Bigram | 0.82 | 0.71 | 0.76 | 555 |
| RF+Unigram | 0.86 | 0.7 | 0.76 | 546 |
| GNB+Trigram | 0.73 | 0.78 | 0.75 | 579 |
| GNB+TFIDF | 0.86 | 0.67 | 0.74 | 541 |
| SVM+TFIDF | 0.91 | 0.68 | 0.74 | 559 |
| GNB+Bigram | 0.77 | 0.71 | 0.73 | 575 |
| MNB+Trigram | 0.66 | 0.83 | 0.73 | 539 |
| RF+TFIDF | 0.83 | 0.64 | 0.69 | 525 |
| DT+TFIDF | 0.67 | 0.68 | 0.67 | 540 |
| SVM+Bigram | 0.71 | 0.64 | 0.67 | 563 |
| KNN+Unigram | 0.75 | 0.62 | 0.66 | 552 |
| MNB+Unigram | 0.75 | 0.6 | 0.66 | 546 |
| RF+Bigram | 0.75 | 0.64 | 0.66 | 550 |
| DT+Unigram | 0.68 | 0.7 | 0.65 | 561 |
| GNB+Unigram | 0.75 | 0.6 | 0.65 | 546 |
| LR+TFIDF | 0.76 | 0.45 | 0.6 | 549 |
| MLP+Trigram | 0.66 | 0.61 | 0.6 | 564 |
| KNN+TFIDF | 0.82 | 0.53 | 0.59 | 544 |
| DT+Bigram | 0.6 | 0.58 | 0.58 | 560 |
| RF+Trigram | 0.69 | 0.57 | 0.55 | 571 |
| MNB+TFIDF | 0.79 | 0.52 | 0.52 | 568 |
| SVM+Trigram | 0.72 | 0.43 | 0.51 | 558 |
| LR+Unigram | 0.75 | 0.46 | 0.5 | 552 |
| KNN+Bigram | 0.69 | 0.45 | 0.48 | 581 |
| KNN+Trigram | 0.53 | 0.51 | 0.41 | 536 |
| DT+Trigram | 0.71 | 0.35 | 0.35 | 551 |
| LR+Bigram | 0.4 | 0.32 | 0.27 | 558 |
| LR+Tri | 0.21 | 0.3 | 0.25 | 568 |

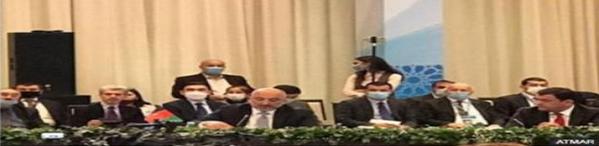

Figure 6 Pashto news article example 1 (https://www.bbc.com/pashto)

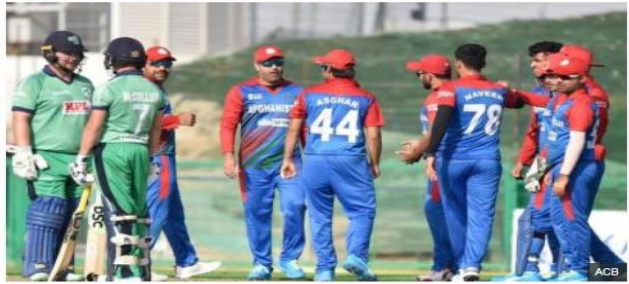

| History | 0 |
| Culture | 0 |
| Economic | 0 |
| Health | 0 |
| Politic | 0 |
| Scientific | 0 |
| Sport | 1 |
| Technology | 0 |
| News | 1 |

Figure 7 Pashto news article example 2 (https://www.bbc.com/pashto)

Table 5 Obtained accuracy related to separate labels in multi-label classification

| Technique | History | Culture | Economic | Health | Politic | Scientific | Sport | Technology | News |
|---|---|---|---|---|---|---|---|---|---|
| DT+Bigram | 0.9 | 0.65 | 0.76 | 0.82 | 0.62 | 0.81 | 0.96 | 0.91 | 0.74 |
| DT+TFIDF | 0.88 | 0.72 | 0.81 | 0.81 | 0.77 | 0.83 | 0.94 | 0.92 | 0.81 |
| DT+Unigram | 0.91 | 0.71 | 0.78 | 0.84 | 0.72 | 0.79 | 0.92 | 0.92 | 0.86 |
| GNB+Bigram | 0.87 | 0.72 | 0.83 | 0.89 | 0.82 | 0.86 | 0.98 | 0.95 | 0.84 |
| GNB+TFIDF | 0.94 | 0.78 | 0.89 | 0.92 | 0.84 | 0.94 | 0.94 | 0.93 | 0.84 |
| GNB+Trigram | 0.89 | 0.73 | 0.82 | 0.87 | 0.8 | 0.83 | 0.96 | 0.92 | 0.9 |
| GNB+Unigram | 0.93 | 0.65 | 0.86 | 0.88 | 0.8 | 0.86 | 0.96 | 0.88 | 0.8 |
| KNN+Bigram | 0.91 | 0.63 | 0.79 | 0.85 | 0.65 | 0.82 | 0.94 | 0.92 | 0.76 |
| KNN+TFIDF | 0.92 | 0.69 | 0.84 | 0.9 | 0.78 | 0.83 | 0.95 | 0.89 | 0.81 |
| KNN+Unigram | 0.93 | 0.72 | 0.83 | 0.88 | 0.8 | 0.88 | 0.96 | 0.9 | 0.85 |
| LR+Bigram | 0.89 | 0.69 | 0.84 | 0.8 | 0.58 | 0.81 | 0.94 | 0.92 | 0.64 |
| LR+TFIDF | 0.91 | 0.7 | 0.84 | 0.8 | 0.8 | 0.86 | 0.95 | 0.9 | 0.9 |
| LR+Unigram | 0.92 | 0.75 | 0.82 | 0.82 | 0.84 | 0.82 | 0.92 | 0.9 | 0.9 |
| MLP+Bigram | 0.95 | 0.78 | 0.89 | 0.89 | 0.79 | 0.9 | 0.96 | 0.96 | 0.88 |
| MLP+TFIDF | 0.96 | 0.86 | 0.92 | 0.93 | 0.89 | 0.92 | 0.96 | 0.95 | 0.94 |
| MLP+Unigram | 0.9 | 0.83 | 0.87 | 0.94 | 0.86 | 0.91 | 0.98 | 0.95 | 0.91 |
| MNB+Bigram | 0.92 | 0.78 | 0.8 | 0.87 | 0.86 | 0.89 | 0.94 | 0.91 | 0.92 |
| MNB+TFIDF | 0.9 | 0.7 | 0.84 | 0.83 | 0.86 | 0.87 | 0.93 | 0.9 | 0.9 |
| MNB+Trigram | 0.89 | 0.73 | 0.8 | 0.83 | 0.76 | 0.87 | 0.92 | 0.88 | 0.9 |
| MNB+Unigram | 0.93 | 0.65 | 0.86 | 0.88 | 0.8 | 0.86 | 0.96 | 0.88 | 0.8 |
| RF+Bigram | 0.921 | 0.733 | 0.835 | 0.898 | 0.733 | 0.858 | 0.976 | 0.909 | 0.851 |
| RF+TFIDF | 0.93 | 0.77 | 0.87 | 0.9 | 0.82 | 0.85 | 0.97 | 0.94 | 0.9 |
| RF+Unigram | 0.93 | 0.78 | 0.86 | 0.93 | 0.85 | 0.92 | 0.99 | 0.92 | 0.89 |
| SVM+Bigram | 0.91 | 0.69 | 0.84 | 0.86 | 0.74 | 0.83 | 0.94 | 0.94 | 0.85 |
| SVM+TFIDF | 0.91 | 0.8 | 0.88 | 0.94 | 0.91 | 0.88 | 0.97 | 0.93 | 0.95 |
| SVM+Trigram | 0.96 | 0.7 | 0.83 | 0.85 | 0.61 | 0.8 | 0.95 | 0.92 | 0.75 |
| SVM+Trigram | 0.96 | 0.7 | 0.83 | 0.85 | 0.61 | 0.8 | 0.95 | 0.92 | 0.75 |
| SVM+Unigram | 0.92 | 0.76 | 0.87 | 0.92 | 0.83 | 0.89 | 0.98 | 0.94 | 0.86 |
| DT+Trigram | 0.93 | 0.61 | 0.81 | 0.85 | 0.6 | 0.81 | 0.97 | 0.92 | 0.68 |
| LR+Tri | 0.95 | 0.68 | 0.82 | 0.81 | 0.58 | 0.77 | 0.9 | 0.9 | 0.67 |
| KNN+Trigram | 0.91 | 0.67 | 0.87 | 0.26 | 0.58 | 0.81 | 0.91 | 0.93 | 0.68 |
| MLP+Trigram | 0.98 | 0.68 | 0.82 | 0.83 | 0.66 | 0.8 | 0.95 | 0.92 | 0.76 |
| RF+Trigram | 0.91 | 0.63 | 0.81 | 0.85 | 0.58 | 0.87 | 0.94 | 0.93 | 0.8 |

## 8.3 Evaluation of the Third Model

We divided the dataset into 80% and 20% portions for train and test sets, respectively in multilabel classifier. By using DistilBERT-base-multilingual-case the obtained accuracy for document multiclass single label classification is 66.31% (table 6). This model achieved 0.68 hamming score and 0.10 hamming loss in Pashto multiclass multilabel classification task (table 4).

The main reason behind low accuracies of pre-trained language models is that we used multilingual base models. It is trained for more than hundred languages. However, Pashto is not in that list. It is difficult for the model to distinguish and recognize Pashto alphabet characters and morphemes. Thus, these models demonstrate cheap performance in Pashto NLP tasks. Therefore, the requirement of a Pashto tokenizer is perceived.

On the other hand, existence of several similar words in some distinct categories results on misclassification of the labels as illustrated in figure 8.

Table 6 Experimental results of the multilabel classification 3rd group models

| Model | Hamming score | Hamming loss |
|---|---|---|
| M- DistilBERT | 0.68 | 0.10 |
| M-BERT | 0.58 | 0.14 |

```
              precision    recall  f1-score   support

     History       0.71      0.75      0.73        16
       Sport       0.50      0.40      0.44        10
  Economical       0.59      0.77      0.67        13
  Technology       1.00      0.86      0.92        14
   Political       0.50      0.43      0.46         7
     Caltural       0.56      0.36      0.43        14
  Scientific       0.56      0.77      0.65        13
      Health       0.88      0.88      0.88         8

    accuracy                           0.66        95
   macro avg       0.66      0.65      0.65        95
weighted avg       0.67      0.66      0.66        95
```

Figure 7 Multiclass classification report using DistilBERT

## 3. CONCLUSIONS

This paper is one of the first state-of-the-art research in Pashto literature text classification analysis. It built the first Pashto documents corpus in two versions one for single and the other for multi-label classification purposes. It also made a lexicon list of Pashto words and developed a multiple classification framework to categorize Pashto documents. This study obtained high accuracy with some classifiers. The highest accuracy achieved by implementing MLP with TFIDF methods in both contexts. The future task is to develop a Pashto tokenizer based on BERT models. Additionally, we will expand our dataset and add a lemmatization task. Moreover, we will observe more state-of-the-art techniques.